\documentclass[10pt,twocolumn,letterpaper]{article}

\usepackage{cvpr}              

\usepackage[accsupp]{axessibility} 
\usepackage[dvipsnames]{xcolor}

\definecolor{cvprblue}{rgb}{0.21,0.49,0.74}
\usepackage[pagebackref,breaklinks,colorlinks,citecolor=cvprblue]{hyperref}
\usepackage{multirow}
\usepackage{wrapfig,lipsum,booktabs}
\usepackage{pifont}
\newcommand{\cmark}{\ding{51}}%

\usepackage{caption}
\def\paperID{12638} 
\def\confName{CVPR}
\def\confYear{2024}

\title{HandDiff: 3D Hand Pose Estimation with Diffusion on Image-Point Cloud}

\author{Wencan Cheng\textsuperscript{1} \quad Hao Tang\textsuperscript{2,3} \quad Luc Van Gool\textsuperscript{2,4} \quad Jong Hwan Ko\textsuperscript{5}\\
\textsuperscript{1}Department of Artificial Intelligence, Sungkyunkwan University\\
\textsuperscript{2}CVL, ETH Zurich \quad
\textsuperscript{3}Carnegie Mellon University \quad
\textsuperscript{4}INSAIT, Sofia Un. St. Kliment Ohridski \\
\textsuperscript{5}College of Information and Communication Engineering, Sungkyunkwan University\\
{\tt\small  \{cwc1260, jhko\}@skku.edu, \{hao.tang, vangool\}@vision.ee.ethz.ch}}

\begin{document}
\maketitle
\begin{abstract}
Extracting keypoint locations from input hand frames, known as 3D hand pose estimation, is a critical task in various human-computer interaction applications. Essentially, the 3D hand pose estimation can be regarded as a 3D point subset generative problem conditioned on input frames. Thanks to the recent significant progress on diffusion-based generative models, hand pose estimation can also benefit from the diffusion model to estimate keypoint locations with high quality. However, directly deploying the existing diffusion models to solve hand pose estimation is non-trivial, since they cannot achieve the complex permutation mapping and precise localization. Based on this motivation, this paper proposes HandDiff, a diffusion-based hand pose estimation model that iteratively denoises accurate hand pose conditioned on hand-shaped image-point clouds. In order to recover keypoint permutation and accurate location, we further introduce joint-wise condition and local detail condition. Experimental results demonstrate that the proposed HandDiff significantly outperforms the existing approaches on four challenging hand pose benchmark datasets. Codes and pre-trained models are publicly available at \url{https://github.com/cwc1260/HandDiff}.
\end{abstract}

\section{Introduction}
\label{sec:intro}
3D hand pose estimation (HPE), which involves estimating the 3D positions of hand keypoints, provides a fundamental conprehension of human hand motion. Therefore, it is essential to facilitate more natural and intuitive interactions between humans and machines and is applicable to various human-computer interaction applications including robotics, gaming, and augmented/virtual reality. In recent years, significant progress has been made in the field of 3D hand pose estimation by applying deep learning techniques and using low-cost depth cameras.

\begin{figure}[!h]
\centering
\includegraphics[width=\linewidth]{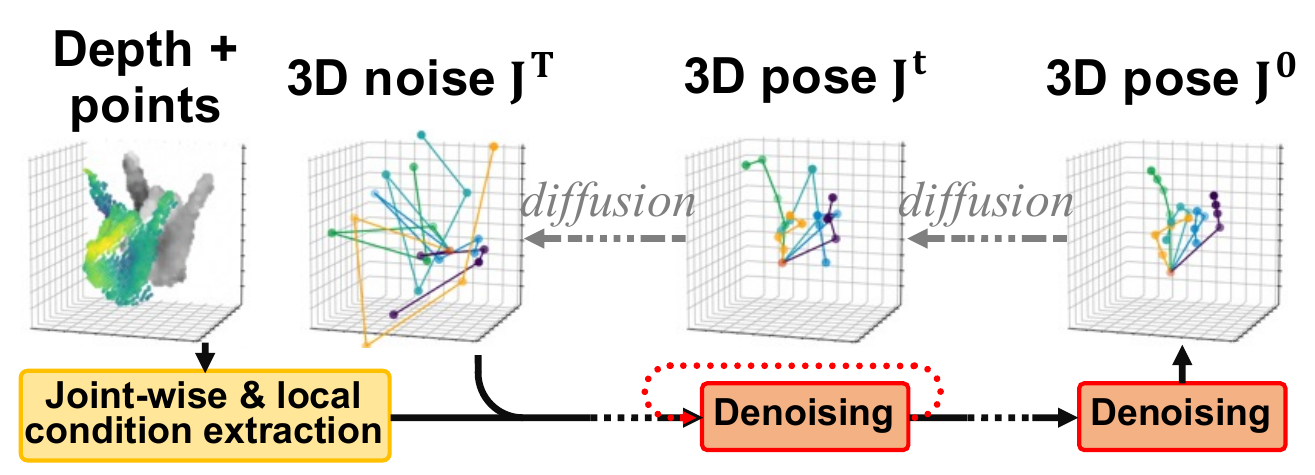}
\caption{Illustration of the hand pose diffusion concept. The model extracts features from input depth images and corresponding point clouds as joint-wise and local conditions to guide the iterative denoising process that recovers accurate hand poses from diffused noisy pose distributions.}
\label{fig:concept}
\end{figure}
Recent developments in 3D Hand Pose Estimation (HPE) based on deep learning \cite{ge2018hand, ge2018point, chen2018shpr, li2019point, cheng2021handfoldingnet, guo2017region, ren2019srn, chen2020pose, fang2020jgr, ren2021pose, du2019crossinfonet, ren2023two}  are primarily divided into two core approaches: regression and detection. While these straightforward solutions have shown notable effectiveness and computational efficiency, these deterministic methods impose limitations on handling ill-posed uncertain cases such as self-occlusions and hand-object occlusions, which are prevalent in real-world hand recognition scenarios. Therefore, in order to ensure the reliability of the estimation, it is imperative to accurately model the uncertainty.

Providentially, a revolutionary approach known as the Diffusion Model (DM) \cite{ho2020denoising, sohl2015deep, luo2021diffusion} has demonstrated remarkable performance in processing uncertainty through modeling of probabilistic distributions. The DM has also exhibited superiority in 3D generative applications including unseen 3D point cloud generation \cite{luo2021diffusion, vahdat2022lion} and uncertain parts completion \cite{lyu2021conditional, zhou20213d}. Therefore, 3D DM can be deployed in 3D HPE as in these 3D conditional generation problems, as both 3D HPE and 3D conditional generation aim to generate a set of keypoints based on a specific condition. More importantly, 3D DM can resolve the ill-posed uncertainty of occlusions by learning the probabilistic distribution of the keypoints.
Based on this inspiration, we apply the diffusion model in generating hand keypoint locations conditioned on the hand depth image/point cloud input, as illustrated in Figure \ref{fig:concept}. To the best of our knowledge, our work is the first attempt to deploy diffusion models in the hand pose estimation task.



While it is theoretically possible to tackle the HPE task using DM models, the direct application of existing 3D DMs \cite{luo2021diffusion, vahdat2022lion} to hand pose estimation tasks is still limited. One of the significant limitations of current 3D DMs is their reliance on a global latent condition, which overlooks crucial local detail information needed for accurate estimation of joint locations. Furthermore, the permutation-equivariant nature of 3D DMs limits their ability to distinguish between joints under simple global conditions, affecting their precision in aligning noisy points with specific target joints.

To fully exploit the potential of the diffusion model in hand pose estimation, we propose HandDiff, a novel approach that incrementally refines the noise distribution to accurately derive a 3D hand pose from multi-modal inputs, including depth images and point clouds. To address inherent limitations in 3D DMs, our model incorporates a joint-wise denoising mechanism that individually denoises various joints during estimation.
Concretely, the proposed model first introduces a joint-wise condition generation module that samples features for each individual joints from both depth image and point cloud. Furthermore, it features a novel local feature-conditioned denoising module, which is the key component to perform the reverse diffusion process. It operates under joint-specific conditions and utilizes local features gathered around the noisy input joint locations. In addition, we propose a novel kinematic correspondence-aware layer that integrates with a graph convolutional operation in order to capture the kinematic relationship between hand joints.


We evaluate HandDiff on four challenging benchmarks, including single-hand ICVL \cite{tang2014latent}, MSRA \cite{sun2015cascaded} and NYU \cite{tompson2014real} dataset, and hand-object DexYCB \cite{chao2021dexycb} datasets. The results show that HandDiff achieves a comparable performance with mean distance errors of 5.76 mm, 6.53 mm, and 7.38 mm on the ICVL, MSRA, and NYU datasets, respectively. The model also significantly outperforms existing state-of-the-art approaches on the DexYCB dataset with the lowest error, achieving the lowest error at 8.06 mm. 

The following is a summary of our primary contributions:
\begin{itemize}[leftmargin=*]
\item We propose a novel diffusion-based model for hand pose estimation that utilizes the depth image and point cloud input as a multi-modal condition. This model progressively denoises a noise distribution, accurately determining the 3D coordinates of hand joints.
\item We propose a novel joint-wise local feature-aware denoising module designed to capture local details surrounding noisy input as a condition for more accurate joint coordinate denoising. Furthermore, this module incorporates a novel kinematic correspondence-aware layer to model the dependencies between joints, thereby enhancing performance. 
\item We perform comprehensive experiments on big and challenging benchmarks that present the new state-of-the-art performance of our proposed method.
\end{itemize}

\section{Related Work}

\noindent\textbf{3D hand pose estimation based on depth image.}
Among the various 3D hand pose estimation approaches that use depth images, conventional 2D CNN-based methods \cite{tompson2014real, ge2016robust, guo2017region, ren2019srn, chen2020pose, fang2020jgr, ren2021pose, du2019crossinfonet} have been widely used due to their simplicity. However, they suffer from limitations such as difficulty in capturing the 3D structure and dependence on the camera's viewpoint.

To overcome these limitations, 3D CNN-based methods \cite{ge20173d,moon2018v2v} were introduced, which use 3D voxelized representations of depth images to capture volumetric information. Although these methods improved the performance of 3D hand pose estimation, they require large amounts of memory and computation, which limits their practical applications.

In contrast, PointNet-based methods \cite{ge2018hand, ge2018point, chen2018shpr, li2019point, cheng2021handfoldingnet} process the point cloud, which is an accurate representation of the 3D structure. PointNet \cite{qi2017pointnet} is a deep learning framework that can handle irregular and unstructured point clouds. The use of PointNet for hand pose estimation was first introduced in HandPointNet \cite{ge2018hand}. 
Point-to-Point model \cite{ge2018point} and SHPR-Net \cite{chen2018shpr} further improved performance by generating the point-wise probability distribution. Subsequently, SHPR-Net \cite{chen2018shpr} combined HandPointNet with an auxiliary semantic segmentation subnetwork to enhance performance. Recently, HandFoldingNet \cite{cheng2021handfoldingnet} introduced a folding concept that reshapes a predefined 2D hand skeleton into hand poses, further improving the estimation accuracy. However, a significant drawback of the point cloud is that querying neighbors from a dense point set for convolution requires heavy computations. Therefore, existing methods commonly use a sparse point cloud, which restricts the performance. 

Hence, in this work, we utilize multi-modal representations that combine 2D depth images and 3D point clouds. Thus, the model is able to efficiently extract dense detail information, as well as effectively capture 3D spatial features for accurate 3D hand pose estimation. Moreover, for the first time, we apply the diffusion model with a PointNet-based denoising process to improve pose estimation performance. 

\noindent\textbf{Diffusion models for pose estimation.}
Diffusion models \cite{ho2020denoising, sohl2015deep}, also known as denoising diffusion probabilistic models (DDPMs), are a family of deep generative models. DM recovers the originally observed data distribution from the perturbed data distribution with gradually injected noise by recurrently denoising the noise of each perturbation step. In recent years, they have seen remarkable success in a variety of computer vision tasks, such as object detection \cite{chen2022diffusiondet}, image synthesis \cite{ho2022cascaded, rombach2022high, saharia2022palette, saharia2022image}, graph generation \cite{niu2020permutation, jo2022score, vignac2022digress}, semantic segmentation \cite{baranchuk2021label, brempong2022denoising}, and pose estimation \cite{shan2023diffusion, gong2022diffpose, holmquist2022diffpose, shan2023diffusion, choi2022diffupose}. 

Existing diffusion-based pose estimation approaches \cite{shan2023diffusion, gong2022diffpose, holmquist2022diffpose, choi2022diffupose} have been used mainly for 3D human pose estimation, which regresses the locations of 3D keypoints of humans from 2D RGB images of the human body. This is because the uncertain 2D-to-3D lifting can be modeled as a probability distribution. Specifically, D3DP \cite{shan2023diffusion} proposed a multi-hypothesis aggregation with joint-wise reprojection to determine the best hypothesis from the diffusion model using the 2D prior. DiffPoses \cite{gong2022diffpose, holmquist2022diffpose} both introduced a heatmap representation of 2D joints to condition the reverse diffusion process. DiffuPose \cite{choi2022diffupose} adopted the graph convolutional network as a denoising function to explicitly learn the connectivity between human joints. In common, these methods all follow the same two-stage scheme, which first requires a trained 2D regression model to obtain 2D keypoints as a prior, and then applies a diffusion model conditioned on the 2D prior to solve the 2D-to-3D lifting problem. Intuitively, the performance of the 2D regression model constrains the denoising quality.

In contrast, our method applies a single-stage denoising process using the conditions from a 3D space. Our method directly accepts raw depth and point cloud frames as conditions in order to take full use of both dense 2D visual features and 3D structural information without requiring any compressed 2D/3D prior information from pre-trained models.
Furthermore, our method leverages the DDIM \cite{song2020denoising} to accelerate inference, since it can complete the thousands of reverse denoising processes in single-digit downsampled timesteps.

\begin{figure*}[t]
\centering
\includegraphics[width=\linewidth]{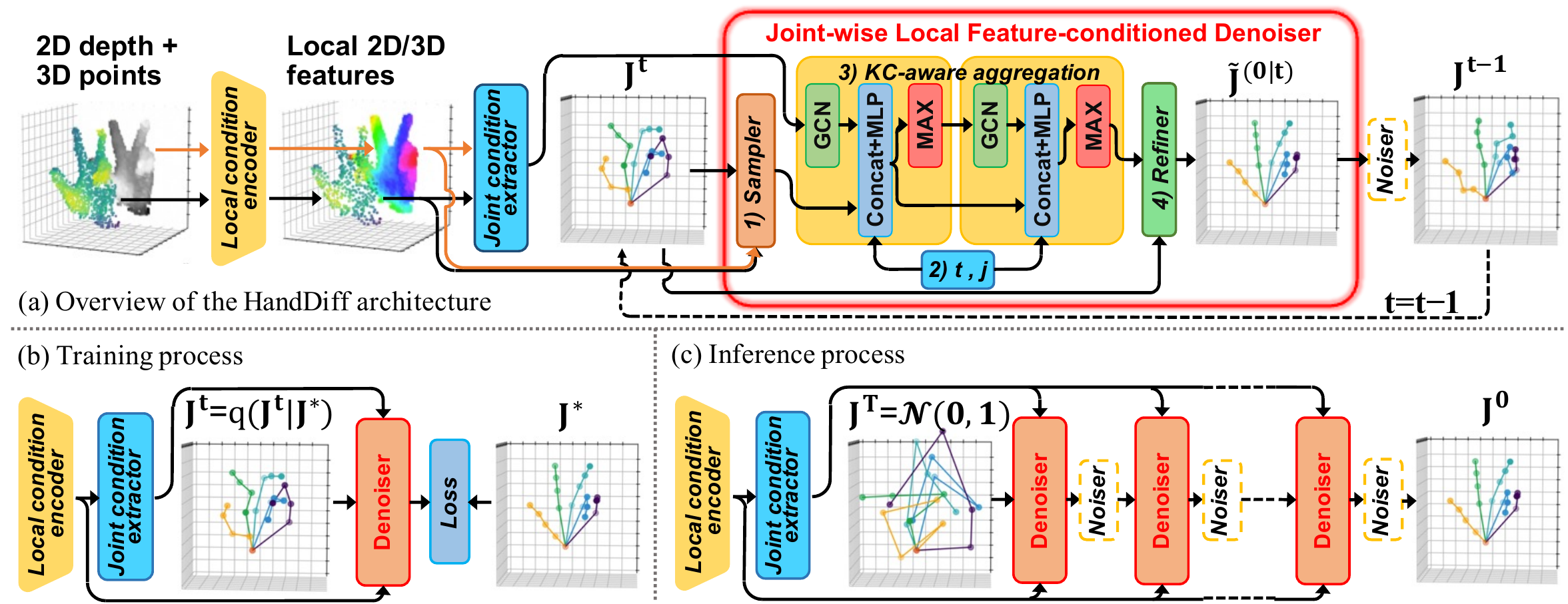}
\caption{The pipeline of the proposed HandDiff. HandDiff takes the normalized point cloud transformed from a 2D depth image as the input. The PointNet-based local condition encoder extracts local features, aka local conditions, from input points. Then, a joint-wise condition extractor aggregates local features into latent features of each joint. Conditioned on the joint-wise conditions and local conditions sampled around each joint, the joint-wise local feature-conditioned denoiser iteratively recovers an accurate 3D hand pose by denoising the diffused noisy pose. Notably, a noiser proposed in DDIM is applied to add noise to the denoised pose for subsequent denoising steps.
}
\label{fig:architecture}
\vspace{-0.4cm}
\end{figure*}


\section{The Proposed Hand Pose Diffusion Model}
HandDiff is a diffusion model that takes a 3D normal distribution and a hand depth image as input and produces the coordinates of the hand joints as output, as shown in Figure~\ref{fig:architecture}. Intuitively, HandDiff iteratively removes noise to refine the joint locations by exploring the local region around each joint conditioned on joint-wise features. The input to HandDiff is a hand depth image $\mathbf{D}_{in} \in \mathbb{R}^{H \times W}$ with a set of sampled 3D point coordinates $\mathbf{P}_{in} \in \mathbb{R}^{N \times 3}$, and the outputs are 3D joint coordinates $\mathbf{J}^0 \in \mathbb{R}^{J \times 3}$ denoised from a randomly initialized normal distribution. The depth image and the $N$ points are first supplied into a local condition encoder that extracts local and global features. We construct a ConvNeXt-based autoenoder to generate a 2D local visual feature map $\mathbf{F}_{2d} \in \mathbb{R}^{H/2 \times W/2 \times d_{2d}}$ and a 2D global vector. Due to the irregularity and disorder of the input point set, we exploit the hierarchical point cloud encoder \cite{qi2017pointnet++, liu2019flownet3d} proposed by PointNet++ \cite{qi2017pointnet++} to extract 3D local geometric features $\mathbf{F}_{3d} \in \mathbb{R}^{N/2 \times d_{3d}}$ and a global 3D vector. Then, the local features are fed into the joint-wise condition extractor to extract joint-wise condition vectors. Afterward, a novel joint-wise local feature-conditioned denoiser is executed $T'$ steps to iteratively denoise joint coordinates by resampling the useful local features around noisy joints and being conditioned by joint-wise conditions.

\subsection{Joint-wise Condition Extraction}

The joint-wise conditions $\mathbf{C} \in \mathbb{R}^{J \times d_{c}}$ essentially are the embeddings that represent the joints in the $d_{c}$-dimensional latent space. Thus, we take the concept of joint-wise embeddings proposed by HandFoldingNet \cite{cheng2021handfoldingnet} to obtain the joint-wise conditions. The generation of the joint-wise conditions (joint-wise embeddings) is sequentially performed by a three-layer bias-induced layer (BIL) \cite{cheng2019point}. 
The concatenation of the 2D and 3D global vectors is replicated $J$ times and fed into the BILs to generate conditions for the $J$ individual joints. The BIL provides joint-wise independent biases that can be regarded as learnable positional embeddings that enable individual mapping for different joints from the same global feature. 

\subsection{Joint-wise Local Feature-conditioned Denoiser}

The denoiser reconstructs an accurate joint coordinate distribution from a noisy distribution under the joint-wise conditions extracted from the hand point cloud. At each time step $t$, the denoiser $\mathcal{D}$ accepts the noisy joint coordinate distribution $\mathbf{J}^{(t)}$, joint-wise condition $\mathbf{C}$, local features $\mathbf{F}_{2d}$, $\mathbf{F}_{3d}$ and time step $t$ as input, and outputs the reconstructed 3D joint coordinates $\widetilde{\mathbf{J}}^{(0|t)}$ without noise:
\begin{equation}
\widetilde{\mathbf{J}}^{(0|t)} = \mathcal{D}(\mathbf{J}^{(t)}, \mathbf{C}, \mathbf{F}_{2d}, \mathbf{F}_{3d}, t).
\end{equation}
The denoiser consists of the following elements: 1) a local feature sampler, 2) a joint indicator \& timestep embedding, 3) a kinematic correspondence-aware aggregation block, and 4) a residual refiner. The sampler samples local features around noisy coordinates $\mathbf{J}^{(t)}$ to form local conditions that contain detailed observations. In order to differentiate between different joints and levels of noise, we introduce a joint indicator and a time-step embedding, respectively. Afterward, the aggregation block fuses the local conditions and joint-wise conditions together and subsequently produces the denoised joint coordinates. In addition, the aggregation also cooperates with graph reasoning to capture kinematic correspondences. In the end, a residual mechanism is applied to refine the noisy joint coordinate distribution to the denoised one.

\noindent\textbf{Local feature sampler.} The sampler collects $K$-nearest-neighbor local 3D points and their corresponding local 3D spatial features around each noisy joint distribution. The sampler also samples $K$-nearest-neighbor 2D pixels and their corresponding local 2D visual features. Note that the 2D pixels are projected onto the same 3D joint space for neighbor querying. The neighbor points and pixels are then translated into a relative coordinate system with the joint location as the origin to eliminate translation variations.

\noindent\textbf{Joint indicator vector \& timestep embedding.} As the denoiser needs to handle different joints with different levels of noise, it has to be explicitly informed of the joint index $j$ and timestep $t$. Following DDPMs, we apply the sinusoidal function on $j$ and $t$ to form a joint indicator vector and time-step embedding. Subsequently, the joint indicator vectors and time-step embeddings are concatenated with each local feature for the subsequent process.

\noindent\textbf{Kinematic correspondence-aware aggregation block.} As visualized in Figure \ref{fig:architecture}, the proposed aggregation block accepts the local features, joint-wise conditions, joint indicator vectors, and timestep embeddings as input, and outputs kinematic correspondence-evolved local features and joint-wise embeddings. Since many recent approaches \cite{fang2020jgr, ren2021pose} suggested that a Graph Convolutional Network (GCN) can effectively model relative kinematic relationships among joints, the joint-wise conditions (embeddings) are first augmented through a GCN, forming the evolved joint-wise conditions: 
\begin{equation}
\mathbf{C'} = ReLU(\mathbf{A}\mathbf{C}\mathbf{W}),
\end{equation}
where $\mathbf{W} \in \mathbb{R}^{d_{c} \times d_{c}}$ is the trainable weights and $\mathbf{A} \in \mathbb{R}^{J \times J \times d_{c}}$ is the channel-wise kinematic correspondence matrix among joints. Afterward, the GCN-evolved joint-wise conditions are concatenated with the joint indicator vectors, timestep embeddings, as well as sampled local features. Subsequently, the concatenated features are sent to a one-layer MLP to generate evolved local features encoding with kinematic correspondence and global prior. Formally, the evolved local feature for the $k$-th local neighbor point of the $j$-th joint is defined as
\begin{equation}
    \mathbf{F'}_{k,j} = MLP([\mathbf{P}_{k,j} - \mathbf{J}_{j}, \mathbf{F}_{k,j}, \mathbf{C'}_j, PE(t), PE(j)]),
\end{equation}
where $\mathbf{P}_{k,j}$ and $\mathbf{F}_{k,j}$ are the coordinate and local feature of the $k$-th local neighbor point of the $j$-th joint, PE is the sinusoidal positional embedding function and `$[\cdot, \cdot]$' is the concatenation operation.

However, the single proposed block is not sufficient for complex denoising. Thus, the proposed block is replicated four times with independent learnable parameters in practice. Furthermore, we introduce a max-pooling layer between every two blocks for providing updated joint-wise embeddings with local information to the latter block:
\begin{equation}
\mathbf{\hat{C}} = MaxPool(\mathbf{F'}).
\end{equation}

\noindent\textbf{Residual refiner.} Similar to the previous 3D generative diffusion model \cite{luo2021diffusion}, the refiner accepts the last joint-wise embeddings $\mathbf{\hat{C}}$ as input to refine the noisy input distribution $\mathbf{J}^{(t)}$. The refiner is a linear transformation with a residual connection. Therefore, the approximated joint coordinates of the current $t$-th timestep are represented as:
\begin{equation}
\widetilde{\mathbf{J}}^{(0|t)}=\mathbf{\hat{C}}\mathbf{W} +\mathbf{J}^{(t)},
\end{equation}
where $\mathbf{W} \in \mathbb{R}^{d_{3} \times 3}$ is the trainable transformation matrix.

\subsection{Training}
HandDiff first corrupts the ground truth joint distribution $q(\mathbf J^{(0)})$ to a noisy distribution $q(\mathbf J^{(t)}|\mathbf J^{(0)})$ by gradually adding noise $\epsilon \sim \mathcal N (\mathbf{0}, \mathbf{I})$ through a forward diffusion Markovian chain, where $t$ is uniformly sampled from the predefined total time steps $T$. Following DDPMs \cite{ho2020denoising}, the forward noise process is formally defined as:
\begin{equation}
\begin{split}
    q(\mathbf J^{(t)}|\mathbf J^{(0)}) = \sqrt{\bar \alpha_t} \mathbf J^{(0)} + \epsilon\sqrt{1 - \bar \alpha_t}, \\
    \text{where } \bar \alpha_t = \prod_{s=0}^t \alpha_t = \prod_{s=0}^t(1-\beta_s).
\end{split}
\end{equation}
Note that $0 < \beta_t < 1$ is the variance of the noise, which is controlled by a linear variance schedule at each time step, as in DDPM \cite{ho2020denoising}.

Subsequently, the noisy joint distribution is supplied to the proposed denoiser to recover the clean joint distribution $\widetilde{\mathbf{J}}^{(0|t)}$, under the joint-wise conditions as well as the local detail conditions. To train the denoiser, $\widetilde{\mathbf{J}}^{(0|t)}$ is under the supervision of the ground truth distribution $\mathbf{J}^*$.

Besides, the joint-wise conditions have to be initialized through training. Therefore, the 3D coordinates $\mathbf{J}^{c}$ linearly transformed from the joint-wise conditions are also under the same supervision.

Following previous regression works \cite{ren2019srn, cheng2021handfoldingnet}, we adopt a smooth L1 loss to supervise training because of its less sensitivity to outliers. The smooth L1 loss is defined as:
\begin{equation}
	L1_{smooth}(\textbf{x}) = \begin{cases}
	50\textbf{x}^2, &|\textbf{x}|<0.01\\
	|\textbf{x}|-0.005, &otherwise
		   \end{cases}
    .
\end{equation}
By using the smooth L1 loss, we supervise the approximated joint distribution by the following joint loss function:
\begin{equation}
\mathcal L = \sum_{j=0}^J L1_{smooth}(\widetilde{\mathbf{J}}^{(0|t)}_{j} - \mathbf{J}_j^*).
\end{equation}

\subsection{Inference}
During inference, a reverse diffusion process is pursued by iteratively applying the denoiser, to recover the uncontaminated joint coordinate distribution. According to recent 2D-to-3D human pose diffusion models \cite{shan2023diffusion, gong2022diffpose, holmquist2022diffpose, choi2022diffupose}, multiple diverse hypotheses for the reverse process can help probabilistic diffusion models to achieve improved accuracy. Our model also samples $H$ initial 3D poses $\mathbf{J}^{(T)}_{0:H}$ from a unit Gaussian distribution. 

Afterward, $H$ pose hypotheses are individually passed to the proposed denoiser to approximate the $H$ uncontaminated joint coordinate distribution $\widetilde{\mathbf{J}}^{(0|t)}_{0:H}$. To obtain the noisy input for the subsequent denoising step $t-1$, we exploit a noiser that adds noise to the denoised distribution following the DDIM \cite{song2020denoising}:
\begin{equation}
p_\theta(\mathbf{J}^{(t-1)}_{0:H}|\widetilde{\mathbf{J}}^{(0|t)}_{0:H}) = \sqrt{\bar \alpha_{t-1}} \widetilde{\mathbf{J}}^{(0|t)}_{0:H} + \sqrt{1 - \bar \alpha_{t-1} - \sigma_{t}^2}\epsilon_t + \sigma_{t}\epsilon,
\end{equation}
where $t$ is started from $T$, $\epsilon_t  = (\mathbf{J}^{(t)}_{0:H} - \sqrt{\bar \alpha_{t}}\widetilde{\mathbf{J}}^{(0)}_{0:H})/\sqrt{1 - \bar \alpha_t}$ is the predicted noise of timestep $t$ and $\sigma_{t} = \sqrt{(1-\bar \alpha_{t-1})(1-\bar \alpha_{t}/\bar \alpha_{t-1})/(1-\bar \alpha_t)}$.

This procedure will be iterated $T'$ times ($T' < T$) to estimate the final denoised distribution $\widetilde{\mathbf{J}}^{(0|t)}_{0:H}$. At the last timestep 0, we average over all hypotheses to aggregate the ultimate uncontaminated joint coordinates:
\begin{equation}
\bar{\mathbf{J}}^{(0)} = {\frac{1}{H}}\sum_{h=0}^H {\widetilde{\mathbf{J}}^{(0)}_{h}}.
\end{equation}

\section{Experiments}
\subsection{Experiment Settings}
\label{sec:set}
We conducted experiments on an NVIDIA TITAN RTX GPU with PyTorch. For training, we used the AdamW optimizer \cite{loshchilov2017decoupled} with beta$_1$ = 0.5, beta$_2$ = 0.999, and learning rate $\alpha$ = 0.001. The input image was resized to 128, the number of input points to the network was randomly sampled to 1,024, the 2D/3D feature depths $d_{2d}$ and $d_{3d}$ are 128, the joint-wise condition depth $d_c$ is 512 and the batch size was set to 64. The diffusion timestep was set to 500 with a cosine variance scheduler. Meanwhile, to avoid overfitting, we adopted online data augmentation with random rotation ([-180.0, 180.0] degrees), 3D scaling ([0.8, 1.2]), and 3D translation ([-20, 20] mm). We trained the model for 30 epochs with a learning rate decay of 0.1 after every 10 epochs.

\subsection{Datasets and Evaluation Metrics}
\noindent
{\bf MSRA Dataset.} The MSRA dataset \cite{sun2015cascaded} provides more than 76K depth image frames, each of which provides $J = 21$ annotated joints, including one joint for the wrist and four joints for each finger. The frames are split into 9 subjects, each of which contains 17 hand gestures. 

\noindent
{\bf ICVL Dataset.} The ICVL dataset \cite{tang2014latent} provides 22K training and 1.6K testing depth frames, each of which provides $J = 16$ annotated joints, including one joint for the palm and three joints for each finger. 

\begin{table}[t!] \small
\caption{Comparison of the proposed method with previous state-of-the-art methods on the ICVL, MSRA, and NYU datasets. Input indicates the input type of 2D depth image (D), 3D voxels (V), or 3D point cloud (P). $\dag$ The results are reported from the retrained VVS following the same cropping strategy \cite{oberweger2017deepprior++} as in the previous state-of-the-art methods \cite{cheng2021handfoldingnet, ge2018hand, ge2018point, sun2015cascaded, ren2021pose, moon2018v2v, ren2021spatial}.}
\centering
\begin{tabular}{c|ccc|c}
\toprule
 \multirow{2}{*}{Method} & \multicolumn{3}{c|}{Mean joint error (mm)}& \multirow{2}{*}{Input}\\
\cline{2-4}
                       & ICVL          &   MSRA        &   NYU     &   \\
\hline
DeepPrior++ \cite{oberweger2017deepprior++}& 8.1          & 9.5           & 12.24           &D  \\
Pose-Ren \cite{chen2020pose}         &6.79           & 8.65         & 11.81         &D   \\
DenseReg \cite{wan2018dense}         &7.3            & 7.2          & 10.2          &D   \\
CrossInfoNet \cite{du2019crossinfonet} &6.73          & 7.86        & 10.08         &D   \\
JGR-P2O \cite{fang2020jgr}           &6.02           & 7.55         & 8.29          &D   \\
SSRN \cite{ren2021spatial}            &6.01           & 7.05         & 7.37         &D   \\
PHG \cite{ren2021pose}                &5.97           & 6.94         & 7.39          &D   \\
VVS \cite{cheng2022efficient} $\dag$             &6.22           & -         & 7.79      &D   \\
\hline
3DCNN \cite{ge20173d}                 & -             &9.6           & 14.1          &V  \\
SHPR-Net \cite{chen2018shpr}         &7.22           & 7.76         & 10.78         &P  \\
HandPointNet \cite{ge2018hand}         &6.94           & 8.5          & 10.54         &P \\
Point-to-Point \cite{ge2018point}    &6.3            &7.7           & 9.10          &P  \\
V2V \cite{moon2018v2v}                 &6.28           &7.59          & 8.42          &V  \\
HandFolding \cite{cheng2021handfoldingnet}   &5.95           &7.34          & 8.58  &P  \\
IPNet \cite{ren2023two}    &5.76          &6.92          & \textbf{7.17}  &D+P  \\
\hline
HandDiff (Ours)    &\textbf{5.72}           &\textbf{6.53}          & 7.38  &D+P  \\
\bottomrule
\end{tabular}
\label{tab:sota}
\end{table}

\begin{table}[t!] \small
\caption{Comparison of the proposed method with previous state-of-the-art methods on the DexYCB datasets.}
\centering
\tabcolsep=3pt
\begin{tabular}{c|cccc|c|c}
\toprule
 \multirow{2}{*}{Method} & \multicolumn{5}{c|}{Mean joint error (mm)}& \multirow{2}{*}{Input}\\
\cline{2-6}
            & S0    &S1     & S2    & S3    & AVG   &  \\
\hline
A2J \cite{xiong2019a2j}        & 23.93 & 25.57 & 27.65 & 24.92 & 25.52 & D\\
Spurr et al. \cite{spurr2020weakly} & 17.34 & 22.26 & 25.49 & 18.44 & 18.44 & RGB\\
METRO    \cite{lin2021end}   & 15.24 &-&-&-&-& RGB \\
Tse et al. \cite{tse2022collaborative} & 16.05 & 21.22 & 27.01 & 17.93 & 20.55 & RGB \\
HandOcc \cite{park2022handoccnet}  & 14.04 &-&-&-&-& RGB \\
IPNet \cite{ren2023two}      & 8.03  & 9.01  & 8.60  & 7.80  & 8.36  & D+P\\

\hline
Ours        & \textbf{7.66}  & \textbf{8.73}  & \textbf{8.40}  & \textbf{7.53} & \textbf{8.07}  & D+P \\
\bottomrule
\end{tabular}
\label{tab:dexycb}
\vspace{-0.4cm}
\end{table}

\begin{figure}
\centering
\includegraphics[width=0.97\linewidth]{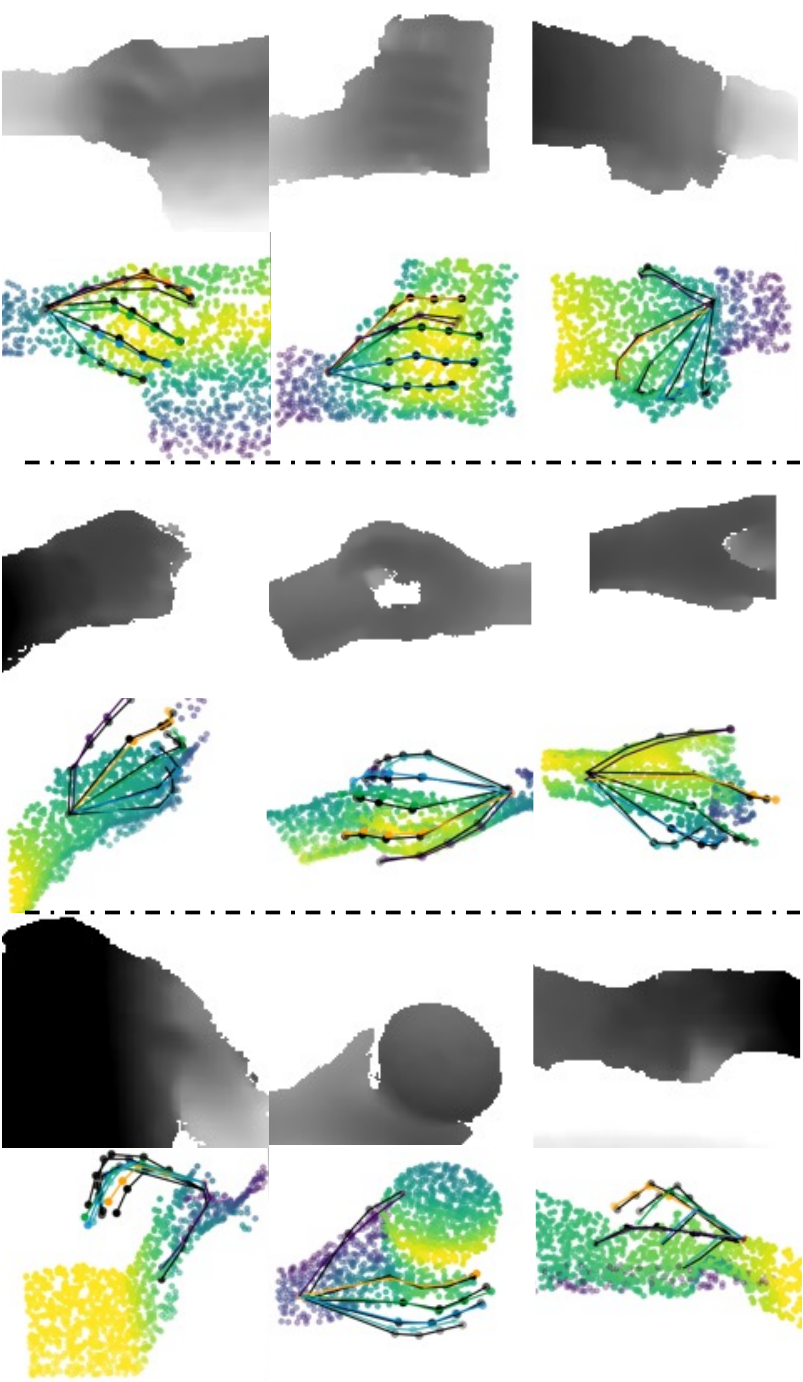}
\caption{Qualitative results of HandDiff on the DexYCB datasets including different grabbing poses (top), self-occlusions (middle), and object occlusions (bottom). Hand-depth images (first rows) are transformed into 3D points (second rows) in order to clearly present occlusions as shown in the figure. Ground truth is shown in black and the estimated joint coordinates of our model are shown in colors.}
\label{fig:ycb}
\vspace{-0.4cm}
\end{figure}

\noindent
{\bf NYU Dataset.} The NYU dataset \cite{tompson2014real} provides depth images captured from three different views by the PrimeSense 3D sensor. Each view contains 72K frames and 8K frames for training and testing, respectively. Following recent works \cite{cheng2021handfoldingnet, ge2018point, ge2018hand}, we use one view for training and testing and selected 14 joints out of a total of 36 annotated joints for evaluation.

\noindent
{\bf DexYCB.} The DexYCB dataset \cite{chao2021dexycb} is a recently released hand-object dataset that consists of 582,000 image frames with 21 annotated joints, 10 different subjects, and 20 YCB objects from 8 camera views. This dataset defines four official dataset split protocols: S0 - seen subjects, camera views, grasped objects; S1 - unseen subjects; S2 - unseen camera views; S3 - unseen grasped objects. 

\noindent
{\bf Evaluation metrics.}
We employ two commonly used metrics, the mean joint error, and the success rate, to evaluate the performance of hand pose estimation. The mean joint error measures the average Euclidean distance between the estimated and ground-truth joint locations for each joint over the testing set. The success rate reveals the percentage of good frames with a mean joint error of less than a certain distance threshold. 

\subsection{Comparison with State-of-the-Art Methods}
\label{sec:sota}

\begin{figure*}
\centering
\includegraphics[width=0.97\linewidth]{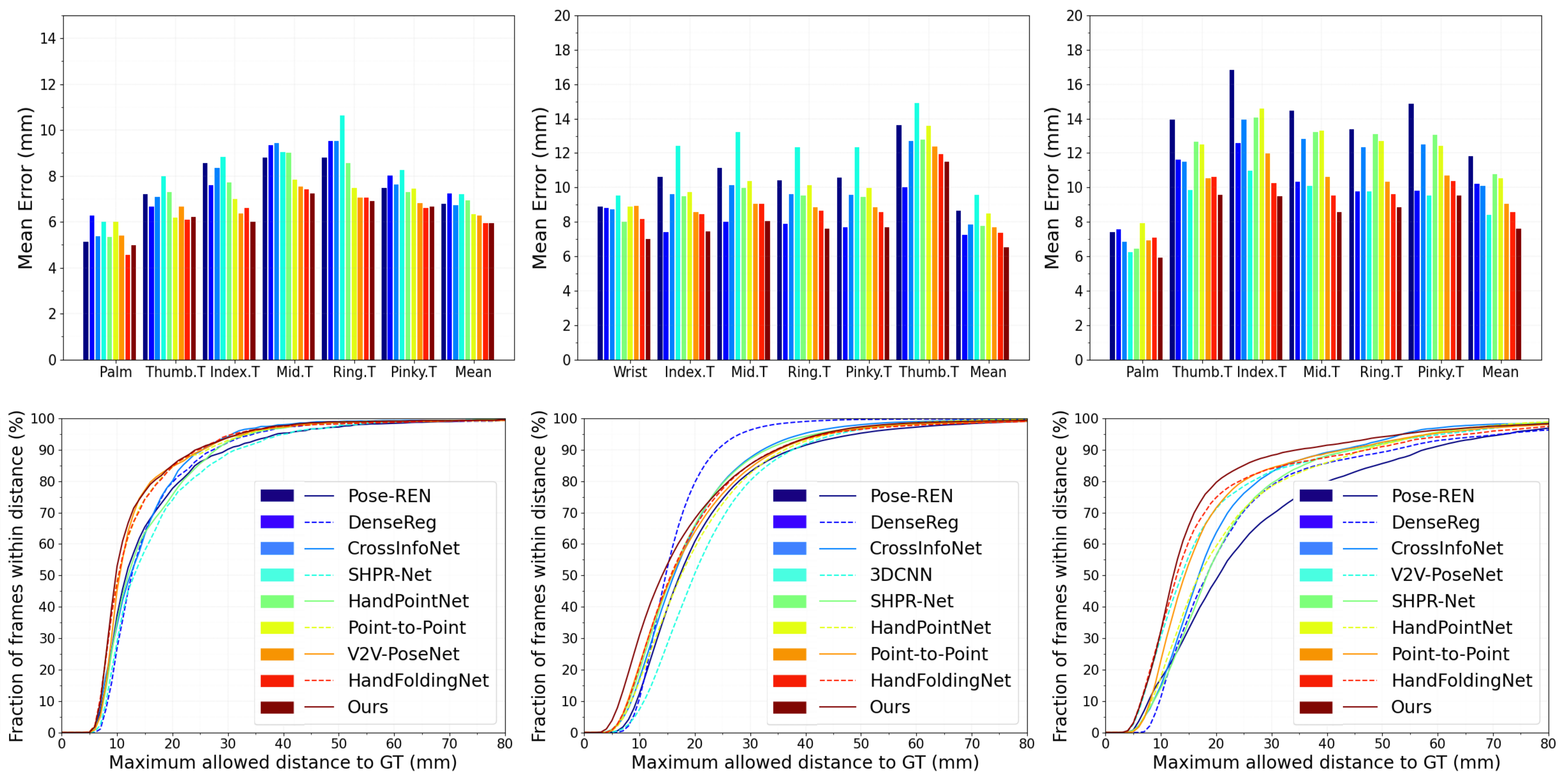}
\caption{Comparison with the state-of-the-art methods using the ICVL (left), MSRA (middle), and NYU (right) dataset. The per joint error (top) and success rate (bottom) are shown in this figure.}
\label{fig:threshold}
\end{figure*}

\begin{figure*}
\centering
\includegraphics[width=0.97\linewidth]{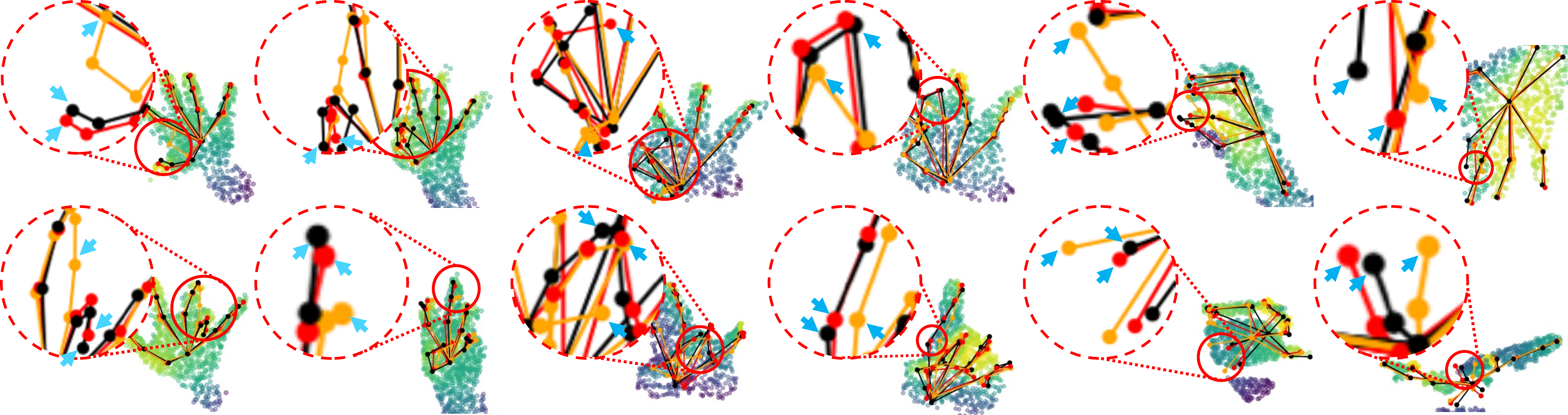}
\caption{Qualitative results of HandDiff on the ICVL (left), MSRA (middle), and NYU (right) datasets. Hand-depth images are transformed into 3D points in order to clearly present occlusions as shown in the figure. Ground truth is shown in black, results from comparative HandFoldingNet \cite{cheng2021handfoldingnet} are shown in orange, and the estimated joint coordinates of our model are shown in red.}
\label{fig:vis}
\end{figure*}

\textbf{Single hand.} We compare HandDiff on the ICVL, MSRA, and NYU dataset with other state-of-the-art methods, including methods with 2D depth images as input: improved DeepPrior (DeepPrior++) \cite{oberweger2017deepprior++}, Pose-Ren \cite{chen2020pose}, dense regression network (DenseReg) \cite{wan2018dense}, CrossInfoNet \cite{du2019crossinfonet}, JGR-P2O \cite{fang2020jgr}, spatial-aware stacked regression network (SSRN)  \cite{ren2021spatial}, pose-guided hierarchical graph network (PHG) \cite{ren2021pose} and virtual view selection (VVS) \cite{cheng2022efficient}, and methods with 3D point cloud or voxels as input: 3DCNN \cite{ge20173d}, SHPR-Net \cite{chen2018shpr}, HandPointNet \cite{ge2018hand}, Point-to-Point \cite{ge2018point}, V2V \cite{moon2018v2v}, HandFoldingNet \cite{cheng2021handfoldingnet} and IPNet \cite{ren2023two}. 

Table \ref{tab:sota} summarizes the results in terms of the mean joint error on the three datasets. The results show that HandDiff achieves the new state-of-the-art record with mean distance errors of 5.72 and 6.53 mm on two challenging datasets, ICVL and MSRA, respectively. The proposed model also achieves the third-lowest error on the NYU dataset. The results also demonstrate that the proposed HandDiff significantly outperforms other 2D image-based methods by large margins since HandDiff directly performs the processing on the 3D space, avoiding the highly non-linear mapping problem of estimating from the 2D image.  Figure \ref{fig:threshold} illustrates that our method significantly outperforms other methods in terms of success rate when the error threshold is lower than 12, 15, and 52 mm on the ICVL, MSRA, and NYU datasets, respectively. 

\noindent \textbf{Hand-object.} We compare HandDiff on the hand-object dataset DexYCB with other state-of-the-art method on the official dataset split protocals, including A2J \cite{xiong2019a2j}, Spurr et al. \cite{spurr2020weakly}, METRO \cite{lin2021end}, Tse et al. \cite{tse2022collaborative}, HandOccNet \cite{park2022handoccnet} and IPNet \cite{ren2023two}. As shown in Table \ref{tab:dexycb}, HandDiff outperforms previous SOTA methods in all four protocols. The qualitative results visualized in Figure \ref{fig:ycb} also reveal that HandDiff can estimate accurate poses from hand-object interaction scenarios with various occlusions.

\subsection{Ablation Study}
\label{sec:ablation}
We conducted extensive ablation experiments to evaluate the contribution of each component proposed in our model. 

\noindent
\textbf{Analysis of different proposed components.} To verify the effectiveness and necessity of the components proposed in this work, we incrementally introduce these components on the existing 3D diffusion probabilistic model (3DDPM) \cite{luo2021diffusion}, which is able to generate complex point cloud conditionally. Briefly, 3DDPM is a share-weight point-wise denoiser conditioned on a global shape latent. We set the number of its output points as the number of hand joints to adapt it to the hand pose estimation task. Afterward, we follow DDIM for the denoising acceleration. Based on this baseline, we incrementally adopt the proposed components and conduct ablations as follows: 1) using local conditions (LC); 2) using joint indicator (JI); 3) using joint-wise condition (JC) and LC; 4) using LC with JI; 5) using JC and LC with JI; 6) using JC and LC with JI and kinematic correspondence (KC); 7) using JC, LC with JI and KC, and multiple hypotheses (MH), which is our full configuration. Note that the use of local conditions without KC is implemented by applying a PointNet layer \cite{qi2017pointnet} on noisy joints. 


\begin{table} \small
\caption{Ablations of different proposed components. All the ablation models are trained and tested on the DexYCB dataset .}
\label{tab:abl}
\small
\centering
\begin{tabular}{ccccc|c}
\toprule
 JC     & LC    & JI    & KC    & MH    & Mean joint error\\
\midrule
        &       &\cmark&       &       & 9.17 mm\\
\hline
        &\cmark&       &       &       & 49.58 mm\\
\hline
\cmark &        &       &       &       & 8.37 mm\\
\hline
\cmark &\cmark&       &       &       & 8.23 mm\\
\hline
        &\cmark&\cmark&       &       & 8.28 mm\\
\hline
\cmark &\cmark&\cmark&       &       & 8.13 mm\\
\hline
        &\cmark&\cmark&\cmark&       & 7.94 mm\\
\hline
\cmark &\cmark&\cmark&\cmark&       & 7.74 mm\\
\hline
\cmark &\cmark&\cmark&\cmark&\cmark& \textbf{7.66} mm\\
\bottomrule
\end{tabular}
\end{table} 

Table \ref{tab:abl} reports the experimental results of the ablations. The results demonstrate that the proposed local condition is permutation-equivariant and thus cannot work solely. The joints must be generated in a specific permutation in order to match the permutation defined by the dataset. Therefore, the proposed joint indicator and joint-wise condition that introduce permutation information are mandatory to improve performance. With the help of the joint indicator and joint-wise condition, the proposed local condition mechanism can significantly reduce the mean joint error by more than 0.9 mm and 0.1 mm, respectively. Furthermore, the proposed kinematic correspondence improves performance by learning the inter-joint relations. Finally, the multiple hypotheses further boost the accuracy.

\begin{table}\small
\caption{Ablations of different modalities of conditions. All the ablation models are trained and tested on the DexYCB dataset .}
\label{tab:modal}
\small
\centering
\begin{tabular}{c|c}
\toprule
Condition modality & Mean joint error (mm)\\
\midrule
2D depth & 8.23 \\
3D points & 9.58 \\
2D depth + 3D points & 7.74 \\
\bottomrule
\end{tabular}

\end{table} 

\noindent
\textbf{Modality of conditions.}  As the quality of conditions determines the quality of pose denoising, we feed the diffusion model with different models of input. Table \ref{tab:modal} shows that the model combining both 2D and 3D conditions presents the optimal estimation performance. The results also reveal that the model with only 2D conditions slightly degrades because of the 3D information loss. On the other hand, the model with only 3D conditions cannot capture dense features from only 1024 points, thus the estimation error significantly increases.

\begin{figure}
\centering
\includegraphics[width=0.8\linewidth]{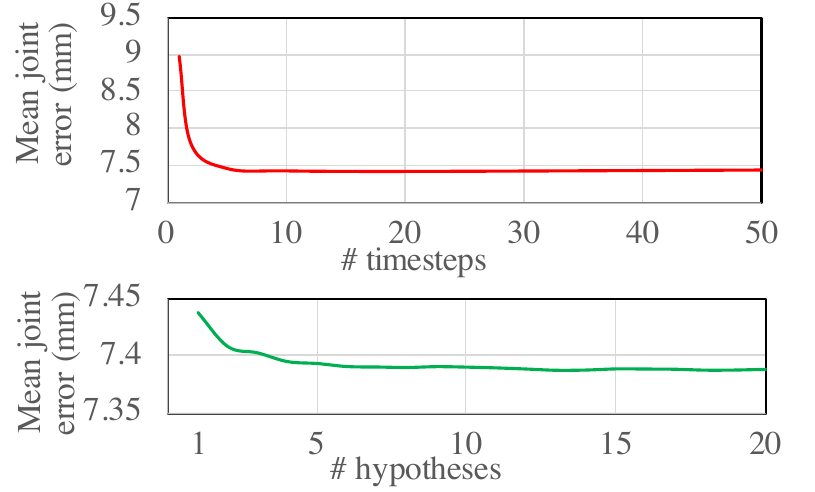}
\caption{Evaluation results of the increasing number of timesteps (top) and hypotheses (bottom) in the denoising process on the NYU dataset. The model is trained with 500 diffusion timesteps.}
\label{fig:time}
\end{figure} 

\noindent
\textbf{Number of denoising timesteps.} As suggested by DDIM \cite{song2020denoising}, the inference process follows a non-Markovian chain. Thus, the number of denoising timesteps can vary for accelerated inference. Figure \ref{fig:time} (top) visualizes the mean joint errors with the increasing number of timesteps during inference (w/o multiple hypotheses). The results show that the model can approach to an acceptable mean joint error with two-step diffusion. The reason is intuitive that the quantity of hand keypoints to be denoised is relatively small compared to other heavy image/point cloud denoising tasks, which normally require hundreds of timesteps. The results also show that 10 timesteps appear as the optimal 7.44 mm. Larger timesteps exhibit a negligible impact on performance. In addition, the computation time and memory of the model are 98 ms and 2.2GB per frame, respectively, for 10 timesteps (1 hypothese). 

\noindent
\textbf{Number of hypotheses.} Figure \ref{fig:time} (bottom) shows the mean joint errors with the different number of hypotheses for denoising. As expected, the error decreases as the hypothese amount increases. However, the improvement becomes marginal when the number is larger than 10.

\section{Conclusion}
\label{sec:conclu}
This paper presented HandDiff, a novel diffusion-based architecture that is capable of reconstructing accurate 3D hand pose iteratively, conditioned on both depth image and point cloud. Experimental results showcased that our network significantly outperforms previous state-of-the-art methods on four challenging datasets. Extensive experiments also reveals the effectiveness of the components proposed in this paper. However, a limitation of HandDiff is its inability to handle scenarios with interacting hands. Future research avenues could explore extensions to bipartite graph learning and skeleton-based analysis to address these limitations and further enhance the model's capabilities.

\noindent
\textbf{Acknowledgement.} This work was supported by the Technology Innovation Program (RS-2023-00235718) funded by the Ministry of Trade, Industry \& Energy (1415187474), and by the Institute of Information and Communication Technology Planning Evaluation (IITP) grants (IITP-2021-0-02068, IITP-2021-0-02052, IITP-2023-2020-0-01821, IITP-2019-0-00421). Wencan Cheng was partly supported by the China Scholarship Council (CSC).



\end{document}